\documentclass[letterpaper]{article} 
\usepackage{aaai2027}  
\usepackage[hyphens]{url}  
\usepackage{graphicx} 
\usepackage{natbib}  
\usepackage{caption} 
\usepackage{algorithm}
\usepackage{algorithmic}
\usepackage{amssymb}
\usepackage{multirow}
\usepackage{makecell}

\usepackage{newfloat}
\usepackage{listings}
\DeclareCaptionStyle{ruled}{labelfont=normalfont,labelsep=colon,strut=off} 
\floatstyle{ruled}
\newfloat{listing}{tb}{lst}{}
\floatname{listing}{Listing}

\usepackage{booktabs}

\nocopyright
\title{PanoSeg3R: Feed-Forward 3D Semantic Segmentation for Panoramic Images with an Automatic Data Curation Pipeline}
\author {
    Heechan Yoon\textsuperscript{\rm 1},
    Dongki Jung\textsuperscript{\rm 1},
    Phuc Nguyen\textsuperscript{\rm 1},
    Ming Lin\textsuperscript{\rm 1},
    Dinesh Manocha\textsuperscript{\rm 1}
}
\affiliations {
    \textsuperscript{\rm 1}University of Maryland, College Park, USA\\
    \{heechany, jdk9405, phucnda, lin, dmanocha\}@umd.edu
}

\begin{document}

\maketitle

\begin{abstract}
We present PanoSeg3R, a feed-forward framework for 3D panoramic semantic segmentation. Unlike existing methods designed for perspective inputs, PanoSeg3R jointly predicts 3D geometry and multi-view semantic segmentation in one single forward pass. Built upon a pretrained reconstruction backbone that supports panoramic images, our approach extends feed-forward 3D reconstruction with a query-based mask decoder. Furthermore, we introduce an automatic panorama data curation pipeline that leverages the complementary strengths of off-the-shelf foundation models to generate reliable pseudo semantic annotations, substantially expanding the training data and improving zero-shot generalization. PanoSeg3R achieves state-of-the-art performance on panoramic 3D semantic segmentation, improving 3D mIoU by up to 16.02 on ScanNet++, while the curated training data further improves zero-shot performance by up to 4.26 and 43.28 mIoU on Stanford2D3D and ToF-360, respectively. Website: \url{https://harryyoon777.github.io/PanoSeg3R/}
\end{abstract}

\section{Introduction}
3D scene understanding is important for many application like immersive media, robotics, and autonomous navigation. Recent learning-based multi-view methods~\cite{li2025iggt, sun2026uni3r, zust2025panst3r} have achieved remarkable progress by enabling accurate 3D geometric reconstruction and dense semantic prediction from RGB observations. However, existing multi-view methods rely on perspective images, whose limited field of view requires capturing numerous overlapping observations to achieve sufficient scene coverage. This acquisition process is often inefficient for large-scale environments~\cite{jung2025im360}.

To address these limitations, omnidirectional cameras are increasingly used because each viewpoint captures richer scene information. Panoramic images are commonly represented using the equirectangular projection (ERP). Semantic segmentation on an ERP image has been explored through various approaches, including distortion-aware network architectures~\cite{li2023sgat4pass, tateno2018distortion, zhang2024behind} and multi-task learning frameworks~\cite{shah2024multipanowise,zhang2026mtpano}. However, existing approaches operate on a single ERP image and therefore cannot produce multi-view consistent segmentation.

\begin{figure}[!t]
    \centering
    \includegraphics[width=\columnwidth]{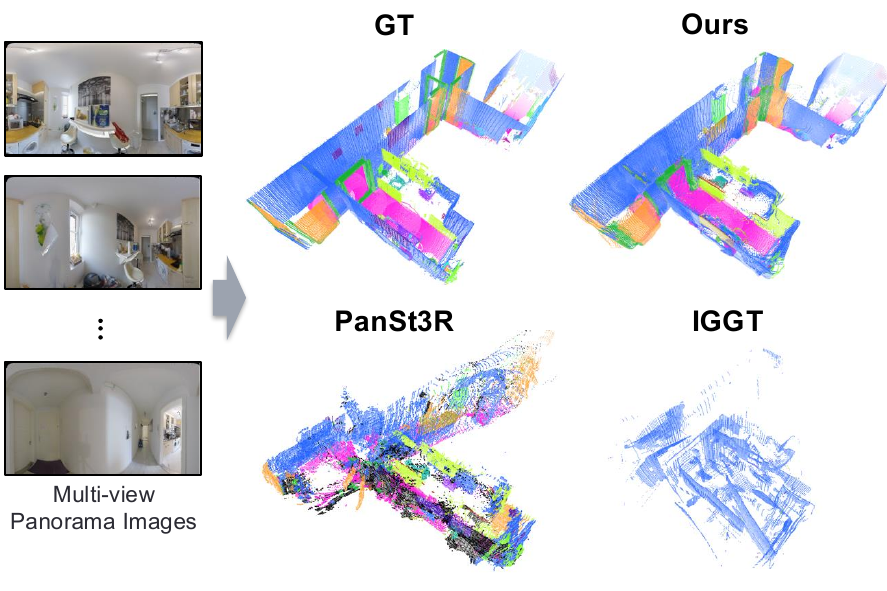}
    \caption{
    PanoSeg3R predicts globally consistent 3D semantic segmentation from multi-view panoramic images, outperforming existing feed-forward methods~\cite{zust2025panst3r, li2025iggt} that rely on cubemap projections.
    }
    \label{fig:teaser}
\end{figure}

Multi-view reconstruction is inherently challenging because predictions from different observations must remain geometrically and semantically consistent. Recent feed-forward methods~\cite{zust2025panst3r,li2025iggt, sun2026uni3r} address this challenge by directly reconstructing 3D scenes from multiple images. However, these methods are primarily designed for distortion-free perspective images and cannot be directly applied to ERP images. To process ERP inputs, panoramas must first be converted into perspective views using cubemap projections~\cite{jiang2021unifuse,wang2020bifuse} or tangent-plane projections~\cite{ai2023hrdfuse,rey2022360monodepth}. These projection-based methods introduce additional computational overhead and compromise geometric and semantic consistency across views, as shown in Figure~\ref{fig:teaser}. Consequently, existing methods still lack a direct and generalizable solution for 3D semantic reconstruction from multi-view ERP images.

Beyond the challenge of developing panorama-native models, the lack of annotated panoramic data poses another major obstacle to advancing semantic segmentation in this domain. Obtaining high-quality annotations for panoramic images is expensive and labor-intensive, making it challenging to construct large-scale annotated panoramic datasets. These highlight the need for scalable approaches to efficiently curate semantic annotations for panoramic images.

\noindent\textbf{Main Results} To address these challenges,
we present PanoSeg3R, a panorama-native feed-forward network for 3D semantic segmentation from multi-view panoramic images. Given a set of ERP images, our model directly predicts dense semantic information in 3D, enabling semantic segmentation to be performed in the native panoramic domain. Built upon a pretrained Wid3R~\cite{jung2026wid3r} backbone, PanoSeg3R introduces a query-based semantic segmentation network. To support training across multiple datasets with different semantic categories, we employ a frozen SigLIP~\cite{tschannen2025siglip} text encoder to represent category names in a shared embedding space. To our best knowledge, PanoSeg3R is the first method to perform 3D semantic segmentation without requiring ERP-to-perspective projection.

In addition, we introduce an automatic pseudo-annotation generation pipeline to alleviate the scarcity of annotated panoramic data. Our pipeline combines complementary foundation models to generate pseudo annotations from unlabeled panoramic images, leveraging class-agnostic segmentation for accurate object boundaries and semantic models to provide category-level predictions. We further employ a vision-language model (VLM)~\cite{bai2025qwen3} to verify and filter the generated annotations, producing reliable semantic annotations for ERP images without manual pixel-wise annotation. This curated data provides additional training supervision and improves the generalization of PanoSeg3R to unseen environments.

We evaluate PanoSeg3R on Matterport3D~\cite{teng2024360bev}, ScanNet++~\cite{yeshwanthliu2023scannetpp}, Stanford2D3D~\cite{armeni2017joint}, and ToF-360~\cite{kanayama2025tof}. PanoSeg3R consistently outperforms existing approaches~\cite{li2025iggt,zust2025panst3r} across both in-domain and zero-shot settings, as shown in Figure~\ref{fig:teaser}. Our contributions are summarized as follows:
\begin{itemize}
    \item We present a feed-forward approach that enables 3D semantic segmentation from multi-view panoramic images.
    \item We develop an automatic semantic panorama data curation pipeline that combines complementary foundation models to generate reliable pseudo annotations from unlabeled ERP images, substantially expanding the training set without manual annotation.
    \item We achieve state-of-the-art performance across four panoramic benchmarks, with up to 16.02 3D mIoU improvement and up to 43.28 mIoU gain from curated data in zero-shot generalization.
\end{itemize}

\begin{figure*}[t]
    \centering
    \includegraphics[width=\textwidth]{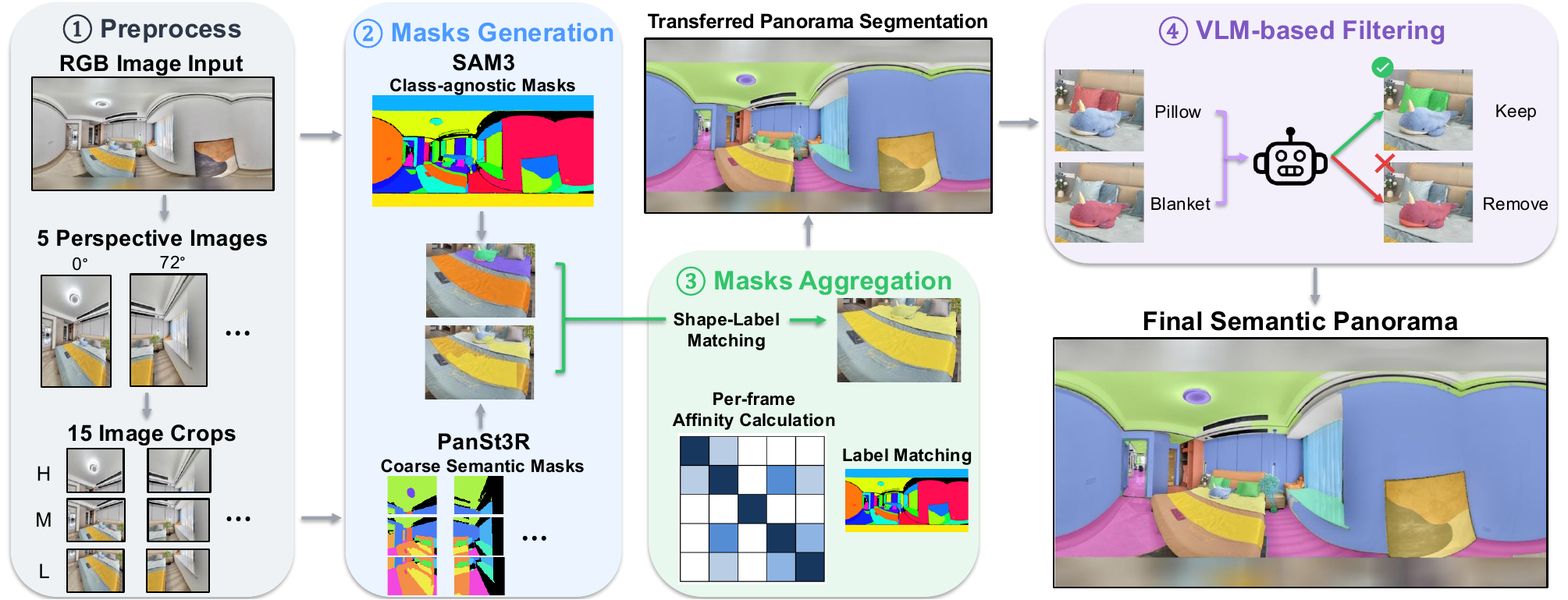}
    \caption{
Overview of the semantic panorama data curation pipeline. Given an RGB ERP image, we first generate overlapping perspective images and then divide them into crops. SAM3~\cite{carion2025sam} generates class-agnostic masks, and PanSt3R~\cite{zust2025panst3r} produces coarse semantic labels from the perspective crops. Their predictions are fused through mask aggregation, followed by VLM-based verification~\cite{bai2025qwen3} to remove inconsistent predictions and produce the final annotations.
    }
    \label{fig:data_curation_pipeline}
\end{figure*}

\section{Related Work}
\subsection{2D Panoramic Semantic Segmentation}
Early approaches to panoramic semantic segmentation explore spherical image representations, including icosahedron~\cite{jiang2019spherical, lee2019spherephd, zhang2019orientation} and tangent images~\cite{eder2020tangent}. Later approaches directly operate on equirectangular projection (ERP) image, introducing specialized convolutional designs such as deformable convolution~\cite{tateno2018distortion} and spherical convolution~\cite{coors2018spherenet, liu2025360}. Transformer-based methods have also been widely explored. SGAT4PASS~\cite{li2023sgat4pass} explicitly models spherical geometry knowledge within the transformer. SFSS-MMSI~\cite{guttikonda2024single} combines the deformable modeling of Trans4PASS+~\cite{zhang2024behind} with Cross Modal Fusion (CMX)~\cite{zhang2023cmx} to leverage complementary modalities for learning more discriminative features. MTPano~\cite{zhang2026mtpano} introduces a multi-task panoramic foundation model with Panorama Dual BridgeNet (PD-BridgeNet), built upon BridgeNet~\cite{zhang2025bridgenet} to improve cross-task consistency. MTPano~\cite{zhang2026mtpano} employs a distortion-aware ERP token mixer for spherical distortions and geometry-aware modulation layers that incorporate absolute position and ray-direction priors. In a different direction, Panosamic~\cite{zhu2026panoramic} leverages a pre-trained SAM~\cite{kirillov2023segment} encoder to incorporate foundation-model knowledge into panoramic semantic segmentation. However, existing methods primarily focus on 2D panoramic semantic segmentation, leaving 3D semantic scene understanding from panoramic images largely unexplored.

\subsection{Joint 3D Reconstruction and Segmentation}
Recent methods have explored jointly modeling 3D reconstruction and semantic understanding within a unified framework. LangSplat~\cite{qin2024langsplat} incorporates vision-language features into 3D Gaussian Splatting~\cite{kerbl20233d}, yet it typically requires per-scene optimization. To enable scalable and generalizable 3D scene understanding across diverse real-world environments, feed-forward models have recently gained increasing attention. Among two-view models, LSM~\cite{fan2024large} unifies semantic and radiance fields to jointly learn 3D geometry, semantic representations, and novel-view synthesis. SIU3R~\cite{wei2026siu3r} further enables native 3D segmentation without relying on 2D-to-3D feature alignment. For multi-view inputs, PanSt3R~\cite{zust2025panst3r} performs 3D reconstruction using MUSt3R~\cite{cabon2025must3r} and feeds features extracted from MUSt3R and DINOv2~\cite{oquab2024dinov2} into Mask2Former~\cite{cheng2022masked} for panoptic segmentation. UNITE~\cite{koch2025unified} adds multiple Dense Prediction Transformer (DPT)~\cite{ranftl2021vision} heads to jointly predict semantic features, instance embeddings, open-vocabulary features, and object articulation within a unified feed-forward framework. Uni3R~\cite{sun2026uni3r} produces a unified 3D Gaussian representation by integrating semantic features extracted from a 2D vision-language model~\cite{li2022language} into a 3D network. IGGT~\cite{li2025iggt} introduces a unified transformer for spatial reconstruction and instance-level understanding, employing 3D-consistent contrastive learning. However, these feed-forward methods are primarily designed for perspective images and do not directly support panoramic inputs. We address this gap by developing a panorama-native framework for 3D semantic segmentation.

\section{Semantic Panorama Data Curation}
To avoid labor-intensive manual annotation of panoramic semantic labels, we propose an automatic data-curation framework that combines the complementary capabilities of PanSt3R~\cite{zust2025panst3r} and SAM3~\cite{carion2025sam}. As illustrated in Figure~\ref{fig:data_curation_pipeline}, the framework consists of four stages: (1) preprocess, (2) mask generation, (3) mask aggregation, and (4) VLM-based filtering. Given an RGB equirectangular projection (ERP) image, we first generate overlapping perspective views and multi-scale image crops for foundation-model inference. PanSt3R produces coarse semantic predictions, while SAM3 provides high-quality class-agnostic object masks with accurate boundaries. During mask aggregation, semantic labels from PanSt3R are transferred to the corresponding SAM3 masks through shape-label matching, after which per-frame affinities and iterative multi-frame merging are used to construct a consistent panoramic segmentation. Finally, a vision-language model (VLM)~\cite{bai2025qwen3} verifies the candidate semantic assignments and removes inconsistent predictions, producing reliable semantic panorama annotations.

\noindent\textbf{Preprocessing.}
Given an RGB ERP image, we first project it into five undistorted perspective images with viewing directions uniformly spaced by $72^\circ$ in azimuth, thereby covering the full $360^\circ$ scene. Each undistorted image is then partitioned into three vertical crops corresponding to its high, middle, and low regions, resulting in fifteen perspective crops in total. Each crop has a resolution of $c_h \times c_w$, with horizontal and vertical overlaps set to $o_h$ and $o_v$, respectively. By reducing the projection distortions inherent in the ERP representation, these perspective crops provide more suitable inputs to PanSt3R for semantic prediction.

\noindent\textbf{Mask Generation.}
We generate complementary segmentation predictions using SAM3 and PanSt3R. SAM3 is applied directly to the input ERP image $\mathbf{I}$ to obtain a set of class-agnostic masks
\begin{equation}
\mathcal{M}^{\mathrm{SAM3}}=\{m_k\}_{k=1}^{N}.
\end{equation}
These masks provide accurate region boundaries but do not contain semantic labels. In parallel, PanSt3R processes the fifteen perspective crops. Its output space is defined by a fixed vocabulary $\mathcal{C}$ containing the 100 predefined ScanNet++ categories~\cite{yeshwanthliu2023scannetpp}. For each crop, PanSt3R predicts a panoptic segmentation map. We discard the instance identifiers and retain the semantic component, producing a set of crop-level regions $(r_i,c_i)$, where $r_i$ is a coarse region mask and $c_i\in\mathcal{C}$ is its predicted semantic label. Thus, SAM3 provides precise boundaries at the ERP image level, while PanSt3R provides coarse semantic labels for the perspective crops.

\noindent\textbf{Mask Aggregation.}
We transfer the PanSt3R labels to the class-agnostic SAM3 masks through shape--label matching. Each crop-level PanSt3R region $r_i$ is first reprojected and splatted back onto the ERP image, producing a ERP-space mask $\widetilde{r}_i$. For every SAM3 mask $m_k$, we compute its affinity with each projected PanSt3R region using intersection-over-union:
\begin{equation}
A_{k,i}={IoU}(m_k,\widetilde{r}_i).
\end{equation}
A pair is considered a valid match when $A_{k,i}>\tau_{\mathrm{IoU}}$. Among all valid matches from the fifteen crops, we assign each SAM3 mask the label of the PanSt3R region with the highest IoU:
\begin{equation}
\hat{c}_k
=
c_{i^*},
\qquad
i^*=\arg\max_i A_{k,i}.
\end{equation}
If no projected PanSt3R region satisfies the threshold, the SAM3 mask remains unlabeled. The resulting semantic masks preserve the accurate boundaries of SAM3, with the semantic labels predicted by PanSt3R transferred to them.

\noindent\textbf{VLM-based Filtering.}
Some transferred labels may still be incorrect due to errors in the coarse PanSt3R predictions or imperfect shape--label matching. To improve annotation reliability, we extract a local RGB crop around each labeled SAM3 mask and highlight the target region in red, as illustrated in Fig.~\ref{fig:data_curation_pipeline}, Step~4. We then prompt a vision-language model (VLM; see Appendix) to determine whether the assigned category correctly describes the highlighted region. Masks whose labels are judged inconsistent are removed, while the verified masks are retained to construct the final semantic panorama annotations.

\section{Method}
\subsection{Panoramic Feature and Geometry Extraction}
We build upon the pretrained Wid3R model~\cite{jung2026wid3r}, a feed-forward 3D reconstruction framework for wide field-of-view cameras. Unlike perspective-based methods that directly predict point maps~\cite{wang2025vggt,wang2025pi}, Wid3R represents each pixel as a camera ray and encodes its direction using spherical harmonics~\cite{piccinelli2025unik3d}, providing a ray-based representation that is robust to panoramic distortions. Given a set of multi-view panoramic images $\{I_i\}_{i=1}^{N}$, Wid3R encodes each panorama into a sequence of image tokens $T_i^{(0)} \in \mathbb{R}^{K \times d_T}$, where $K$ denotes the number of tokens per panorama and $d_T$ denotes the token dimension. The token sequences are then processed by a unified transformer with alternating frame and global attention layers, yielding representations $T_i^{(l)}$ after the $l$-th transformer layer, where $1 \leq l \leq L$. These representations capture both local visual details within each panorama and scene-level context across multiple views. For each panorama $I_i$, Wid3R predicts a dense ray-direction map $\mathbf{R}_i$ and estimates the corresponding radial-distance map $\mathbf{D}_i$. The resulting 3D point map is computed as $\mathbf{P}_i = \mathbf{R}_i \odot \mathbf{D}_i$, where $\odot$ denotes element-wise multiplication.

\subsection{Semantic Mask Prediction and Classification}
Inspired by~\cite{zust2025panst3r, cheng2022masked}, we propose a set-prediction framework for panoramic multi-view semantic segmentation.

\noindent\textbf{Query Refinement.} We introduce a set of $N_q$ learnable semantic queries, $\{Q_n^{(0)}\}_{n=1}^{N_q} \in \mathbb{R}^{d_Q}$, shared across all input views. These queries are fed into a decoder consisting of alternating cross-attention with the final-layer tokens $T^{(L)}$ and self-attention layers, producing refined semantic queries, $\hat{Q}_n = \mathrm{Dec}(Q_n^{(0)}, T^{(L)})$.

\noindent\textbf{Semantic Classification.} To support training across multiple datasets with different semantic categories, we use a frozen SigLIP~\cite{tschannen2025siglip} text encoder. The class names of $J$ semantic categories are encoded into text embeddings $\{t_j\}_{j=1}^{J}$. Each refined semantic query $\hat{Q}_n$ is projected into the text embedding space through a linear projection head, $q_n^{\mathrm{cls}} = \mathrm{Lin}_{\mathrm{cls}}(\hat{Q}_n)$. The semantic classification scores are computed as the cosine similarity between the projected query embedding and each text embedding:
\begin{equation}
    p_{n,j} = \mathrm{sim}(q_n^{\mathrm{cls}}, t_j).
\end{equation}

\noindent\textbf{Semantic Mask Prediction.} For semantic mask prediction, we employ a Dense Prediction Transformer (DPT)~\cite{ranftl2021vision} head to aggregate the multi-scale token representations $\{T^{(l)}\}_{l=1}^{L}$ and decode them into a dense feature map, $F_i \in \mathbb{R}^{d_F \times \frac{H}{2} \times \frac{W}{2}}$, for each panoramic view $I_i$.

\noindent\textbf{Low-level Feature Injection.} Compared with perspective images, each patch token in an equirectangular projection (ERP) image covers a substantially larger field of view, causing fine-grained spatial details to be lost during patch tokenization. Consequently, the decoded features become less descriptive of the underlying image content. To compensate for this information loss, we introduce an image stem that extracts complementary low-level features from the input ERP image and injects them into the decoded features. The image stem consists of a strided convolution, residual convolutional blocks, and a $1 \times 1$ projection, producing $S_i = \mathcal{S}(I_i) \in \mathbb{R}^{d_F \times \frac{H}{2} \times \frac{W}{2}}$. To preserve the horizontal continuity of the ERP image, circular padding is applied across all convolutional layers. The extracted stem features $S_i$ are fused with the feature map $F_i$, and then processed through a convolutional projection head $\mathcal{H}(\cdot)$ to produce the final dense mask feature map $\tilde{F}_i \in \mathbb{R}^{d_M \times \frac{H}{2} \times \frac{W}{2}}$:
\begin{equation}
    \tilde{F}_i = \mathcal{H}\left(F_i + S_i\right).
\end{equation}
Finally, each refined query $\hat{Q}_n$ is projected to a mask embedding via a multi-layer perceptron, $q_n^{\mathrm{mask}} = \mathrm{MLP}_{\mathrm{mask}}(\hat{Q}_n)$. The semantic mask for the $n$-th query on view $I_i$ is computed via a dot product with the fused mask feature map:
\begin{equation}
    M_{n,i} = \mathrm{Sigmoid}\left( (q_n^{\mathrm{mask}})^{\top} \tilde{F}_i \right) \in \mathbb{R}^{\frac{H}{2} \times \frac{W}{2}}.
\end{equation}

\begin{figure}[t]
    \centering
    \includegraphics[width=\columnwidth]{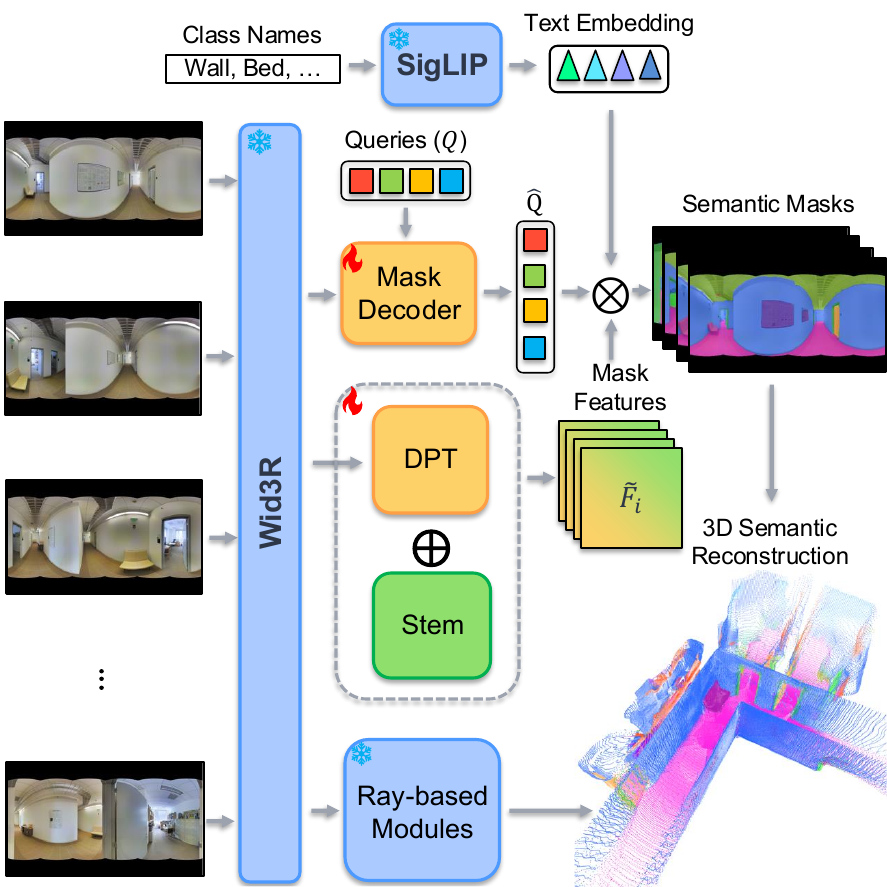}
    \caption{
    Model architecture. Multi-view ERP images are processed by the pretrained Wid3R backbone to extract features. These features are refined by a query-based decoder, while a DPT head produces dense feature maps. An image stem injects low-level features into the feature maps to complement the decoded representations. We reconstruct 3D geometry through the ray-based modules and project the predicted semantic masks to obtain consistent 3D semantic segmentation.
    }
    \label{fig:architecture}
\end{figure}

\noindent\textbf{Training Loss.}
Following Mask2Former~\cite{cheng2022masked}, we optimize the network using three loss terms:
\begin{equation}
\mathcal{L}
=
\lambda_c \mathcal{L}_{cls}
+
\lambda_d \mathcal{L}_{dice}
+
\lambda_b \mathcal{L}_{bce},
\end{equation}
where $\mathcal{L}_{cls}, \mathcal{L}_{dice}$ and $\mathcal{L}_{bce}$ denote classification, Dice, and binary cross-entropy losses, respectively.

\begin{table}[t]
    \centering
    \footnotesize
    \setlength{\tabcolsep}{4pt}
    \resizebox{\columnwidth}{!}{
    \begin{tabular}{llcrrc}
    \toprule
    & \textbf{Dataset} & \textbf{GT Semantic} & \textbf{Scenes} & \textbf{Images} & \textbf{Classes} \\
    \midrule
    \multirow{4}{*}{Train}
    & ScanNet++ & \checkmark & 856 & 4,670 & 100 \\
    & Matterport3D & \checkmark & 61 & 7,829 & 20 \\
    & Realsee3D & $\times$ & 1,000 & 24,263 & 100 \\
    & OmniScenes & $\times$ & 28 & 2,995 & 100 \\
    \midrule\midrule
    \multirow{4}{*}{Eval}
    & ScanNet++ & \checkmark & 50 & 243 & 100 \\
    & Matterport3D & \checkmark & 18 & 2,014 & 20 \\
    & Stanford2D3D & \checkmark & 6 & 1,413 & 13 \\
    & ToF-360 & \checkmark & 4 & 179 & 39 \\
    \bottomrule
    \end{tabular}
    }
    \caption{
    Statistics of training and evaluation datasets. ``GT Semantic'' denotes official ground-truth annotations. Datasets lacking GT use pseudo-labels generated via our semantic panorama data curation pipeline following 100 ScanNet++ categories~\cite{yeshwanthliu2023scannetpp} to boost performance.
    }
    \label{tab:datasets}
\end{table}

\section{Experiments}

\begin{figure*}[t]
    \centering
    \includegraphics[width=\textwidth]{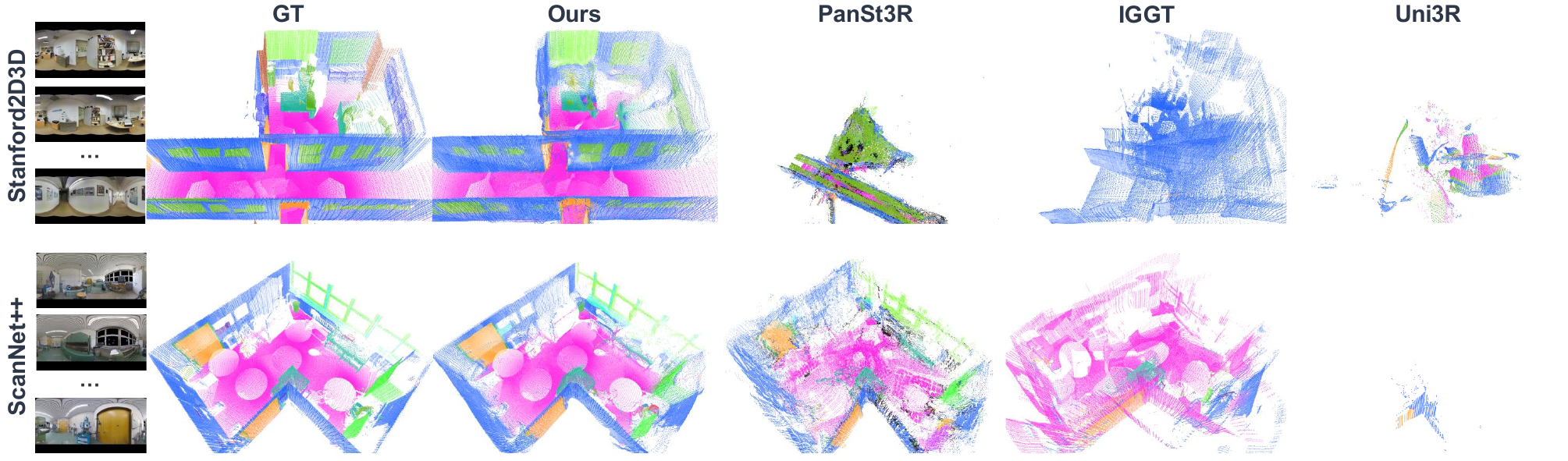}
    \caption{Qualitative results on ScanNet++~\cite{yeshwanthliu2023scannetpp} and Stanford2D3D~\cite{armeni2017joint}. Our method demonstrates superior 3D geometric reconstruction and semantic segmentation performance on panoramic images compared to existing approaches.}
    \label{fig:qualitative_comparison1}
\end{figure*}

\begin{figure}[!t]
    \centering
    \includegraphics[width=\columnwidth]{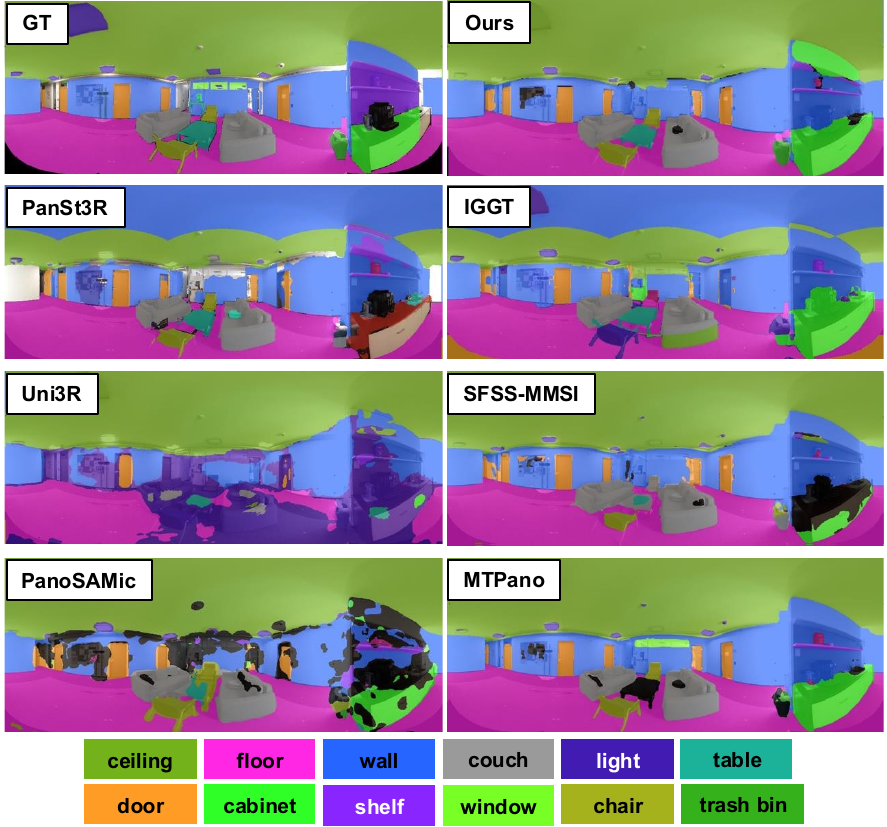}
    \caption{
    Qualitative evaluation of 2D semantic segmentation on ToF-360~\cite{kanayama2025tof}.
    }
    \label{fig:qualitative_comparison2}
\end{figure}

\subsection{Datasets and Implementation Details}
We train our model on a diverse collection of indoor multi-view panorama datasets, including ScanNet++~\cite{yeshwanthliu2023scannetpp}, Matterport3D~\cite{teng2024360bev}, Realsee3D~\cite{Li2025realsee3d_data}, and OmniScenes~\cite{kim2021piccolo}. For Realsee3D, we use only the 1,000 real-world scenes and exclude the synthetic subset. The semantic annotations for Realsee3D and OmniScenes are generated using the proposed semantic panorama data curation pipeline, while the official annotations are used for ScanNet++ and Matterport3D. Detailed statistics of the training datasets can be found in Table~\ref{tab:datasets}.

For our training strategy, we adopt the sampling and augmentation procedures from Wid3R~\cite{jung2026wid3r}. During training, multi-view images are sampled from the same scene using a distance-based sampling strategy that favors spatially nearby viewpoints to ensure sufficient visual overlap. Furthermore, Geometric transformations are applied consistently to both the RGB images and the corresponding ground-truth semantic masks, while photometric augmentations are applied only to the RGB images.

We freeze the pretrained Wid3R~\cite{jung2026wid3r} backbone during training, optimizing only the semantic segmentation head. The network is optimized using the AdamW~\cite{loshchilov2017fixing} optimizer with a peak learning rate of $1 \times 10^{-4}$, $\lambda_c=0.5$, $\lambda_d=1$, and $\lambda_b=1$ on RTX A5000 GPUs. Input equirectangular projection (ERP) images are resized to $518\times336$. For the data curation pipeline, ERP images of size $1024 \times 512$ are converted into fifteen perspective crops with $c_w=256$, $c_h=192$, $o_h = 85$, and $o_v = 64$, with $\tau_{\mathrm{IoU}} = 0.7$ applied in the mask aggregation stage.

\subsection{Experiment Settings}
We compare our method with panorama-specific methods and perspective-based baselines. The latter include PanSt3R~\cite{zust2025panst3r}, IGGT~\cite{li2025iggt}, and Uni3R~\cite{sun2026uni3r} which are feed-forward 3D reconstruction models with segmentation capabilities and, primarily designed for perspective inputs. For these baselines, we apply the cubemap projection to convert panoramic inputs into the perspective format required by the models. The predicted results are then reprojected to the ERP space for evaluation.

For 2D evaluation, semantic predictions are resized to $1024\times512$ before computing the mean Intersection-over-Union (mIoU) and mean class accuracy (mAcc).

For 3D evaluation, covisible panoramas within each scene are jointly processed to obtain a unified point cloud with per-point semantic labels. For perspective-based baselines, the corresponding cubemap views from all covisible panoramas are used as input. We follow the evaluation protocol of IGGT~\cite{li2025iggt}. Specifically, the predicted point cloud is aligned to the ground-truth point cloud using an Umeyama~\cite{umeyama1991least} transformation followed by Iterative Closest Point (ICP) refinement. Both point clouds are then discretized into a voxel grid with a voxel size of 0.1m. A prediction is considered correct only if it matches both the spatial occupancy and semantic category of the corresponding ground-truth voxel. We report the resulting 3D mIoU and mAcc.

\subsection{Experimental Results}

\begin{table}[t]
\centering
\small
\setlength{\tabcolsep}{4pt}
\resizebox{\columnwidth}{!}{
\begin{tabular}{lccc|cccc}
\toprule
\multirow[c]{3}{*}[-0.6ex]{Method} &
\multirow[c]{3}{*}[-0.6ex]{\makecell[c]{Input\\Modalities}} &
\multicolumn{2}{c|}{Matterport3D} &
\multicolumn{4}{c}{ScanNet++} \\

\cmidrule(lr){3-4}
\cmidrule(lr){5-8}

&
&
\multicolumn{2}{c|}{2D} &
\multicolumn{2}{c}{2D} &
\multicolumn{2}{c}{3D} \\

\cmidrule(lr){3-4}
\cmidrule(lr){5-6}
\cmidrule(l){7-8}

&
&
mIoU$\uparrow$ & mAcc$\uparrow$ &
mIoU$\uparrow$ & mAcc$\uparrow$ &
mIoU$\uparrow$ & mAcc$\uparrow$ \\

\midrule

\multicolumn{8}{l}{\textbf{Single Panorama Input}} \\

\midrule

PanoSAMic & \multirow[c]{2}{*}{RGB-D}
& 45.22 & 61.09
& 26.69 & 51.86
& -- & -- \\

SFSS-MMSI &
& 49.85 & 61.41
& 33.86 & 42.90
& -- & -- \\

\cmidrule(lr){1-8}

PanoSAMic & \multirow[c]{3}{*}{RGB}
& 44.33 & 63.12
& 26.74 & 54.27
& -- & -- \\

SFSS-MMSI &
& 44.03 & 55.10
& 30.82 & 39.82
& -- & -- \\

MTPano &
& 53.84 & 70.27
& -- & --
& -- & -- \\

\midrule

\multicolumn{8}{l}{\textbf{Multi View Input}} \\

\midrule

PanSt3R$^{\dagger}$ & \multirow[c]{4}{*}{RGB}
& -- & --
& 27.14 & 34.39
& 9.26 & 12.12 \\

IGGT$^{\dagger}$ &
& -- & --
& 7.63 & 13.14
& 2.30 & 3.26 \\

Uni3R$^{\dagger}$ &
& -- & --
& 5.61 & 7.31
& 1.26 & 1.78 \\

Ours &
& 49.77 & 62.39
& 26.73 & 34.48
& \textbf{17.28} & \textbf{22.56} \\

\bottomrule
\end{tabular}
}
\caption{
Comparison of 2D and scene-level 3D semantic segmentation performance on Matterport3D~\cite{teng2024360bev} and ScanNet++~\cite{yeshwanthliu2023scannetpp}.
$^{\dagger}$ denotes perspective foundation models evaluated on panoramic inputs via six cubemap faces.
Our method achieves competitive 2D segmentation performance while substantially outperforming perspective-based approaches in 3D semantic segmentation.
}
\label{tab:comparison_in_domain}
\end{table}

\noindent\textbf{2D Semantic Segmentation.}
Table~\ref{tab:comparison_in_domain} presents the 2D semantic segmentation results on Matterport3D~\cite{teng2024360bev} and ScanNet++~\cite{yeshwanthliu2023scannetpp}. We compare our method with 2D panorama-specific approaches, including PanoSAMic~\cite{chamseddine2026panosamic} and SFSS-MMSI~\cite{guttikonda2024single}, as well as the recent panorama foundation model MTPano~\cite{zhang2026mtpano}. We further compare with perspective-based 3D reconstruction models, including PanSt3R~\cite{zust2025panst3r}, IGGT~\cite{li2025iggt}, and Uni3R~\cite{sun2026uni3r} which are evaluated on panoramic inputs through cubemap projection. Qualitative results are presented in Fig. ~\ref{fig:qualitative_comparison2}.

Our method achieves competitive performance with existing panorama-specific approaches on both datasets. While some panorama-specific methods obtain higher 2D segmentation scores, these methods are primarily optimized for 2D single view semantic prediction. In contrast, our framework is designed to preserve semantic consistency across panoramic observations and further enables scene-level 3D semantic segmentation.

Compared with perspective-based approaches on ScanNet++, our method outperforms IGGT and Uni3R by 19.10 and 21.12 mIoU, respectively, while remaining competitive with PanSt3R. These results highlight the efficacy of native panoramic modeling without relying on perspective decomposition.

\noindent\textbf{3D Semantic Segmentation.}
Table~\ref{tab:comparison_in_domain} reports multi-view 3D semantic segmentation results on ScanNet++. Our method outperforms all baselines, surpassing PanSt3R, IGGT, and Uni3R by 8.02, 14.98, and 16.02 mIoU, respectively. These results demonstrate the advantage
of directly modeling panoramic inputs for multi-view 3D semantic segmentation. Qualitative comparisons are shown in Fig.~\ref{fig:qualitative_comparison1}.

\noindent\textbf{Zero-shot Evaluation}
\begin{table}[t]
\centering
\small
\setlength{\tabcolsep}{4pt}
\resizebox{\columnwidth}{!}{
\begin{tabular}{lccccc|cc}
\toprule
\multirow[c]{3}{*}[-0.6ex]{Method} &
\multirow[c]{3}{*}[-0.6ex]{\makecell[c]{Input\\Modalities}} &
\multicolumn{4}{c|}{Stanford2D3D} &
\multicolumn{2}{c}{ToF-360} \\

\cmidrule(lr){3-6}
\cmidrule(l){7-8}

&
&
\multicolumn{2}{c}{2D} &
\multicolumn{2}{c|}{3D} &
\multicolumn{2}{c}{2D} \\

\cmidrule(lr){3-4}
\cmidrule(lr){5-6}
\cmidrule(l){7-8}

&
&
mIoU$\uparrow$ & mAcc$\uparrow$ &
mIoU$\uparrow$ & mAcc$\uparrow$ &
mIoU$\uparrow$ & mAcc$\uparrow$ \\

\midrule

\multicolumn{8}{l}{\textbf{Single Panorama Input}} \\

\midrule

PanoSAMic & \multirow[c]{3}{*}{RGB}
& 55.31 & 63.54 & -- & --
& 37.99 & 57.26 \\

SFSS-MMSI &
& 54.30 & 60.54 & -- & --
& 44.98 & 52.14 \\

MTPano &
& \textbf{65.61} & 68.92 & -- & --
& \textbf{61.05} & \textbf{71.04} \\

\midrule

\multicolumn{8}{l}{\textbf{Multi View Input}} \\

\midrule

PanSt3R$^{\dagger}$ & \multirow[c]{4}{*}{RGB}
& 54.31 & 57.15 & 12.65 & 14.13
& 29.59 & 33.37 \\

IGGT$^{\dagger}$ &
& 27.38 & 41.41 & 4.02 & 4.71
& 18.90 & 27.19 \\

Uni3R$^{\dagger}$ &
& 36.56 & 40.51 & 7.36 & 9.05
& 25.01 & 26.76 \\

Ours &
& \textbf{65.15} & \textbf{70.97} & \textbf{34.83} & \textbf{39.01}
& \textbf{56.51} & \textbf{64.42} \\

\bottomrule
\end{tabular}
}
\caption{
Zero-shot 2D and 3D semantic segmentation on Stanford2D3D~\cite{armeni2017joint} and ToF-360~\cite{kanayama2025tof}. ToF-360 is evaluated only in 2D since camera poses are unavailable for 3D reconstruction. $^{\dagger}$ denotes perspective foundation models evaluated on panoramic inputs via six cubemap faces.
}
\label{tab:comparison_zeroshot}
\end{table}
Table~\ref{tab:comparison_zeroshot} presents zero-shot results on Stanford2D3D~\cite{armeni2017joint} and ToF-360~\cite{kanayama2025tof}. For fair cross-dataset evaluation, we consider only the semantic categories shared between the training and target datasets. Our method achieves competitive 2D performance with the recent panorama foundation model MTPano while consistently outperforming existing panorama-specific methods on Stanford2D3D. In multi-view 3D semantic segmentation, our method achieves 35.44 mIoU, surpassing PanSt3R, Uni3R, and IGGT by 22.79, 28.08, and 31.42 mIoU, respectively, demonstrating strong cross dataset generalization.

\subsection{Ablation Study}
\begin{table}[t]
\centering
\small
\setlength{\tabcolsep}{4pt}
\resizebox{\columnwidth}{!}{
\begin{tabular}{l|cccc|cc}
\toprule
\multirow{3}{*}[-0.6ex]{Setting}
& \multicolumn{4}{c|}{Stanford2D3D}
& \multicolumn{2}{c}{ToF-360} \\
\cmidrule(lr){2-5}\cmidrule(l){6-7}
& \multicolumn{2}{c}{2D}
& \multicolumn{2}{c|}{3D}
& \multicolumn{2}{c}{2D} \\
\cmidrule(lr){2-3}\cmidrule(lr){4-5}\cmidrule(l){6-7}
& mIoU$\uparrow$ & mAcc$\uparrow$
& mIoU$\uparrow$ & mAcc$\uparrow$
& mIoU$\uparrow$ & mAcc$\uparrow$ \\
\midrule

\textbf{(a) Training Data} & & & & & & \\
Base
& 60.89 & 67.87
& 33.79 & 38.10
& 13.23 & 20.87 \\

\quad + OmniScenes
& 61.66 & 67.84
& 34.22 & 38.28
& 37.24 & 43.16 \\

\quad + RealSee3D
& 60.84 & 68.61
& 34.09 & 38.48
& 51.94 & 61.36 \\

\quad + Both
& \textbf{65.15} & \textbf{70.97}
& \textbf{34.83} & \textbf{39.01}
& \textbf{56.51} & \textbf{64.42} \\

\midrule

\textbf{(b) Image Stem} & & & & & & \\
Without
& 59.90 & 68.02
& \textbf{34.85} & \textbf{39.28}
& 49.89 & 57.59 \\

With (Ours)
& \textbf{65.15} & \textbf{70.97}
& 34.83 & 39.01
& \textbf{56.51} & \textbf{64.42} \\

\bottomrule
\end{tabular}
}
\caption{Ablation study on zero-shot evaluation on Stanford2D3D~\cite{armeni2017joint} and ToF-360~\cite{kanayama2025tof}. We analyze the effects of the proposed panorama data curation pipeline and the image stem. The curated pseudo semantic datasets substantially improve cross-dataset generalization, with the image stem also contributing to the overall performance.}
\label{tab:ablation_pseudo}
\end{table}

\begin{table}[t]
\centering
\small
\setlength{\tabcolsep}{2pt}
\resizebox{\columnwidth}{!}{
\begin{tabular}{lccc|ccc}
\toprule
\multirow[c]{2}{*}[-1.5ex]{Method} &
\multicolumn{3}{c|}{Stanford2D3D} &
\multicolumn{3}{c}{ToF-360} \\

\cmidrule(lr){2-4}
\cmidrule(lr){5-7}

&
mIoU$\uparrow$ &
\makecell[c]{Wrong\\removed}$\uparrow$ &
\makecell[c]{Correct\\removed}$\downarrow$ &
mIoU$\uparrow$ &
\makecell[c]{Wrong\\removed}$\uparrow$ &
\makecell[c]{Correct\\removed}$\downarrow$ \\

\midrule

\multicolumn{7}{l}{\textbf{Baseline}} \\

\midrule

SAM3 + SigLIP
& 15.22 & -- & --
& 19.18 & -- & -- \\

SAM3 + text prompts
& 47.76 & -- & --
& 29.15 & -- & -- \\

\midrule

\multicolumn{7}{l}{\textbf{Ours}} \\

\midrule

PanSt3R raw
& 53.85 & -- & --
& 42.77 & -- & -- \\

+ SAM3
& \textbf{55.26} & -- & --
& \textbf{43.81} & -- & -- \\

+ VLM
& 55.08 & \makecell{24/122 \\ (19.7\%)} & \makecell{39/755 \\ (5.2\%)}
& 43.41 & \makecell{77/290 \\ (26.6\%)} & \makecell{21/586 \\ (3.6\%)} \\

\bottomrule
\end{tabular}
}
\caption{
Ablation study of our data curation pipeline, including pseudo-annotation quality. For VLM filtering evaluation, we report correctly removed mislabeled masks over total mislabeled masks and mistakenly removed correct masks over total correct masks, together with their respective percentages.
}
\label{tab:ablation_curation}
\end{table}

Table~\ref{tab:ablation_pseudo}(a) validates our data curation pipeline on zero shot benchmarks Stanford2D3D and ToF360. Combining both curated datasets achieves the best performance across all benchmarks, boosting Stanford2D3D 2D and 3D mIoU by 4.26 and 1.04, and ToF360 2D mIoU by 43.28. This demonstrates that the two curated datasets provide complementary semantic and scene diversity for cross-dataset generalization. We further analyze the effect of the proposed image stem. As shown in Table~\ref{tab:ablation_pseudo}(b), the image stem consistently improves 2D semantic segmentation while preserving comparable 3D performance, as 3D mIoU is primarily influenced by 3D scene geometry rather than fine boundary details.

We next evaluate the effectiveness of each stage in the proposed data curation pipeline. To directly assess pseudo-annotation quality, we compare the curated pseudo annotations against ground-truth ones, as shown in Table~\ref{tab:ablation_curation}. We consider two standalone 2D baselines. The first uses SAM3~\cite{carion2025sam} with text prompts. The second uses SAM3 to generate class agnostic masks, cropping each extracted instance and predicting its class label using SigLIP~\cite{tschannen2025siglip}. Compared to these 2D baselines, raw PanSt3R predictions achieve superior performance. Subsequent refinement of these raw PanSt3R predictions using SAM3 further boosts mIoU by 1.41 and 1.04 on Stanford2D3D and ToF360, demonstrating that SAM3 effectively refines coarse boundaries. Finally, regarding vision-language model (VLM) filtering, we report the number of removed masks over the total number of mislabeled masks and the number of mistakenly removed masks over the total number of correctly labeled masks, followed by their respective percentages. The VLM~\cite{bai2025qwen3} removes 19.7\% and 26.6\% of mislabeled masks on Stanford2D3D and ToF-360, respectively, while mistakenly removing 5.2\% and 3.6\% of correctly labeled masks. This explains why VLM filtering can lead to a lower mIoU. Since the VLM can only remove masks, it reduces False Positives (FPs) but cannot recover False Negatives (FNs). Moreover, incorrectly removing valid masks converts True Positives (TPs) into FNs, which can outweigh the benefit of FP reduction.

\section{Conclusion, Limitations and Future Work}
We presented PanoSeg3R, a feed-forward network for 3D semantic segmentation directly from multi-view panoramic images. We demonstrate that our method achieves superior performance in 3D reconstruction and semantic segmentation, which are challenging for existing perspective-based models. In addition, we introduce an automatic data curation pipeline that generates reliable pseudo-annotations from unlabeled panoramic images. This approach substantially expands the training data and further improves the zero-shot performance of PanoSeg3R. One limitation of our approach is that the current data curation pipeline is primarily limited to indoor environments, as its category set is defined based on an indoor scene dataset. Moreover, the pipeline currently focuses on semantic annotations and does not support instance-level or panoptic annotations. In future work, we plan to extend the pipeline to outdoor environments and develop multi-view-consistent data curation methods for instance and panoptic segmentation.

\bibliography{aaai2027}

\newpage


\appendix

\section*{Appendix}

In this supplementary material, we first provide additional details of our semantic panorama data curation pipeline, including visualization examples, implementation details, and ablation studies. Second, we describe the detailed architecture and ablation study of the proposed image stem. Finally, we provide additional evaluation details and experimental results, including the shared semantic categories used for zero-shot evaluation, efficiency comparisons, failure cases, and additional qualitative comparisons.

\section{Semantic Panorama Data Curation}
\begin{figure*}[t]
    \centering
    \includegraphics[width=\textwidth]{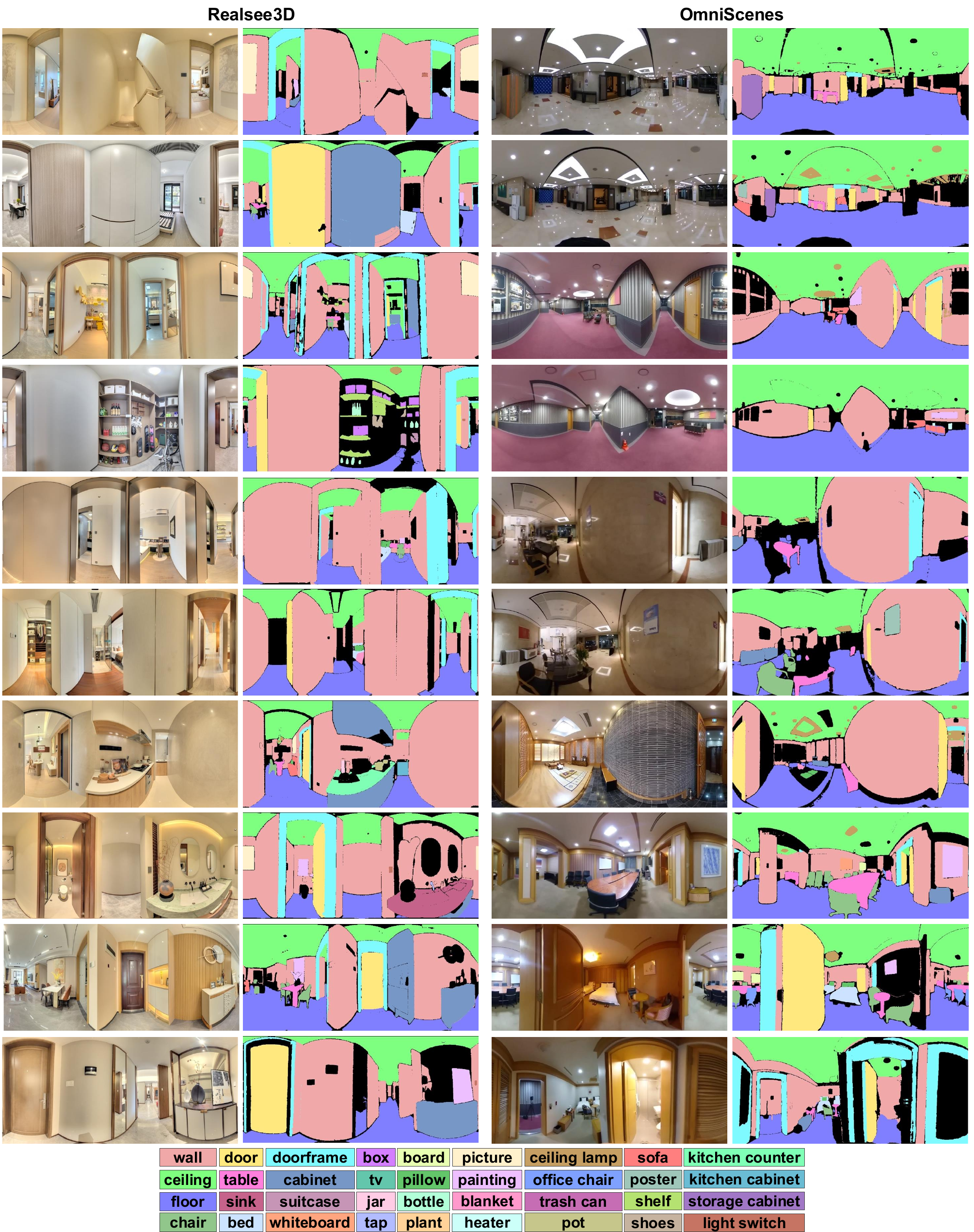}
    \caption
    {
Visualization of the semantic annotations generated by our data curation pipeline. Each example consists of an input RGB ERP image and its corresponding semantic annotation.
    }
    \label{fig:supp_qualitative_curation}
\end{figure*}

Figure~\ref{fig:supp_qualitative_curation} shows qualitative examples of the semantic annotations generated by our data curation pipeline on the Realsee3D~\cite{Li2025realsee3d_data} and OmniScenes~\cite{kim2021piccolo} datasets.

\subsection{Mask Generation.}
We conduct an ablation study on crop configurations to investigate their effects on the PanSt3R~\cite{zust2025panst3r} predictions used in Stage 2 of our curation pipeline. In Table~\ref{tab:ablation_crop}, we evaluate these configurations on ScanNet++~\cite{yeshwanthliu2023scannetpp}, Stanford2D3D~\cite{armeni2017joint}, and ToF-360~\cite{kanayama2025tof}. We vary the number of crops, crop size $c_w \times c_h$, and horizontal and vertical overlap $o_w$ and $o_h$. The overlap is adjusted for each configuration to provide sufficient overlap between adjacent crops while maintaining coverage of the full equirectangular projection (ERP) image. This overlap helps PanSt3R produce consistent predictions across adjacent crops and reduces label discrepancies near crop boundaries.
The 5$\times$3 configuration achieves the best performance on Stanford2D3D and ToF-360. Specifically, it outperforms the 4$\times$3 setup by up to 2.84 mean Intersection-over-Union (mIoU) and exceeds other variants in the 5 $\times$ series by up to 9.03. Although the 5$\times$5 configuration performs best on ScanNet++, increasing the number of vertical crops generally degrades performance, as observed in the 4 $\times$ series. We therefore adopt 5$\times$3 as the default configuration for its strong and consistent performance across datasets.

\begin{figure}[!t]
    \centering
    \includegraphics[width=\columnwidth]{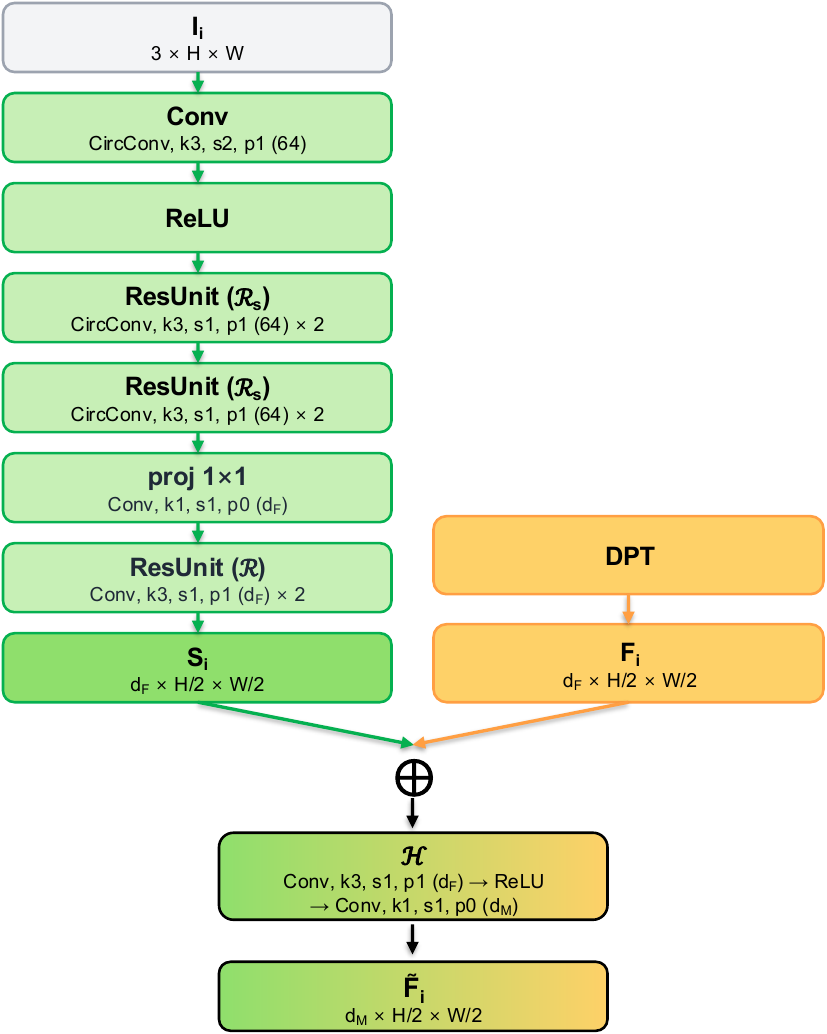}
    \caption{
    Details of the image stem. Conv is a convolutional layer and CircConv is a convolutional layer with circular padding, k is the kernel size, s is the stride, and p is the padding. The numbers in the parentheses denote the number of filters. $\oplus$ denotes element wise addition.
    }
    \label{fig:image_stem}
\end{figure}

\begin{table*}[t]
\centering
\small
\setlength{\tabcolsep}{10pt}
\resizebox{\textwidth}{!}{
\begin{tabular}{cccc|cc|cc|cc}
\toprule

\multirow[c]{2}{*}[-2.0ex]{\#Crops} &
\multirow[c]{2}{*}[-2.0ex]{$c_w \times c_h$} &
\multirow[c]{2}{*}[-2.0ex]{$o_w$} &
\multirow[c]{2}{*}[-2.0ex]{$o_h$} &

\multicolumn{2}{c|}{ScanNet++} &
\multicolumn{2}{c|}{Stanford2D3D} &
\multicolumn{2}{c}{ToF-360} \\

\cmidrule(lr){5-6}
\cmidrule(lr){7-8}
\cmidrule(lr){9-10}

&
&
&
&
\multicolumn{2}{c|}{2D} &
\multicolumn{2}{c|}{2D} &
\multicolumn{2}{c}{2D} \\

\cmidrule(lr){5-6}
\cmidrule(lr){7-8}
\cmidrule(lr){9-10}

&
&
&
&
mIoU$\uparrow$ &
mAcc$\uparrow$ &
mIoU$\uparrow$ &
mAcc$\uparrow$ &
mIoU$\uparrow$ &
mAcc$\uparrow$ \\

\midrule

8 (4$\times$2)
& $256\times256$
& 82
& 96
& 29.37 & 36.28
& 52.52 & 55.13
& 39.32 & 43.47 \\

12 (4$\times$3)
& $256\times192$
& 82
& 64
& 29.00 & 35.66
& 51.01 & 53.51
& 40.03 & 44.10 \\

16 (4$\times$4)
& $256\times160$
& 82
& 64
& 26.66 & 33.88
& 48.79 & 51.70
& 37.41 & 41.92 \\

20 (4$\times$5)
& $256\times144$
& 82
& 64
& 22.36 & 29.27
& 45.22 & 48.98
& 33.82 & 38.48 \\

\midrule

10 (5$\times$2)
& $256\times256$
& 85
& 96
& 20.99 & 28.85
& 44.82 & 49.88
& 34.09 & 39.42 \\

\textbf{15 (5$\times$3) (Ours)}
& $256\times192$
& 85
& 64
& 29.72 & 37.10
& \textbf{53.85} & \textbf{56.03}
& \textbf{42.77} & \textbf{45.92} \\

20 (5$\times$4)
& $256\times160$
& 85
& 64
& 29.74 & 36.89
& 51.81 & 54.24
& 41.60 & 44.97 \\

25 (5$\times$5)
& $256\times144$
& 85
& 64
& \textbf{29.82} & \textbf{37.38}
& 52.70 & 54.87
& 41.64 & 44.89 \\

\bottomrule
\end{tabular}
}
\caption{
Ablation study of crop configurations on ScanNet++~\cite{yeshwanthliu2023scannetpp}, Stanford2D3D~\cite{armeni2017joint}, and ToF-360~\cite{kanayama2025tof}. We vary the number of crops, crop size $c_w \times c_h$, and horizontal and vertical overlap $o_w$ and $o_h$ to evaluate their effects on the PanSt3R~\cite{zust2025panst3r} predictions used in our curation pipeline. \textbf{15 (5$\times$3)} denotes the default configuration used in our experiments.
}
\label{tab:ablation_crop}
\end{table*}

\begin{table}[t]
\centering
\small
\setlength{\tabcolsep}{2pt}
\resizebox{\columnwidth}{!}{
\begin{tabular}{lccc|cc|cc|cc}
\toprule
\multirow[c]{3}{*}[-0.6ex]{Setting} &
\multirow[c]{3}{*}[-0.6ex]{$\#\mathcal{R}_s$} &
\multirow[c]{3}{*}[-0.6ex]{Pad} &
\multirow[c]{3}{*}[-0.6ex]{$\mathcal{R}$} &
\multicolumn{4}{c|}{Stanford2D3D} &
\multicolumn{2}{c}{ToF-360} \\

\cmidrule(lr){5-8}
\cmidrule(l){9-10}

& & & &
\multicolumn{2}{c|}{2D} &
\multicolumn{2}{c|}{3D} &
\multicolumn{2}{c}{2D} \\

\cmidrule(lr){5-6}
\cmidrule(lr){7-8}
\cmidrule(l){9-10}

& & & &
mIoU$\uparrow$ & mAcc$\uparrow$ &
mIoU$\uparrow$ & mAcc$\uparrow$ &
mIoU$\uparrow$ & mAcc$\uparrow$ \\

\midrule

A (w/o stem)
& -- & -- & --
& 59.90 & 68.02
& 34.85 & 39.28
& 49.89 & 57.59 \\

\midrule

B
& 1 & zero & O
& 58.64 & 66.22
& 34.37 & 39.02
& 53.36 & 60.88 \\

C
& 1 & circ & O
& 63.86 & 70.28
& 34.35 & 39.29
& 54.72 & 60.98 \\

D
& 1 & circ & X
& 60.48 & 70.13
& 34.24 & 38.82
& 56.02 & \textbf{64.45} \\

\midrule

E
& 2 & zero & O
& 63.80 & 68.95
& 33.72 & 37.97
& 52.37 & 56.98 \\

\textbf{F (w/ Stem)}
& 2 & circ & O
& \textbf{65.15} & \textbf{70.97}
& 34.83 & 39.01
& \textbf{56.51} & 64.42 \\

G
& 2 & circ & X
& 63.50 & 69.42
& \textbf{35.05} & \textbf{39.36}
& 45.49 & 51.44 \\

\bottomrule
\end{tabular}
}
\caption
{
Ablation study of the image stem on Stanford2D3D~\cite{armeni2017joint} and ToF-360~\cite{kanayama2025tof}. We vary the number of residual convolutional blocks $\#\mathcal{R}_s$, the padding strategy used in $\mathcal{R}_s$, and the use of the final residual convolutional unit $\mathcal{R}$. Setting A denotes the baseline without the image stem, while Setting F denotes the full configuration used in our experiments.
}
\label{tab:ablation_image_stem}
\end{table}

\subsection{VLM Filtering.}
We employ Qwen3-VL-8B~\cite{bai2025qwen3} to identify and remove masks with incorrect semantic labels through chain-of-thought prompting. For each labeled mask, we extract a local RGB region surrounding the mask and highlight its contour in red. The VLM receives the resulting image crop along with the following prompt:

\begin{quote}
\small
\ttfamily
Look at the region outlined in RED in this indoor room photo.

Think step by step:

1. Identify the object primarily enclosed by the outlined region.

2. Determine whether "\{lbl\}" is a reasonable semantic label for that object based on its appearance and the surrounding scene.

3. Answer "Correct" or "Incorrect".
\end{quote}
Here, \texttt{\{lbl\}} denotes the semantic label assigned to the corresponding mask. We parse the response and retain masks classified as Correct while removing those classified as Incorrect.

\section{Image Stem}
Figure~\ref{fig:image_stem} illustrates the detailed architecture of the proposed image stem. The overall pipeline of our mask feature extraction consists of the image stem, Dense Prediction Transformer (DPT)~\cite{ranftl2021vision}, and the projection head $\mathcal{H}$. Green and orange blocks represent the image stem and DPT, respectively. The stem comprises of a strided convolution, residual convolutional blocks $\mathcal{R}_s$, and a $1\times1$ projection. We further denote the final residual convolutional unit by $\mathcal{R}$. The stem feature $S_i$ and the DPT decoder feature $F_i$ are fused and fed into the projection head $\mathcal{H}$ to yield the dense mask feature $\tilde{F}_i$. This feature is subsequently used to derive semantic masks following Equation 6 of the main paper.

The results in Table~\ref{tab:ablation_image_stem} show that the image stem improves performance over the baseline on both Stanford2D3D~\cite{armeni2017joint} and ToF-360~\cite{kanayama2025tof}. To analyze the choice of padding, we compare standard zero padding with circular padding applied within $\mathcal{R}_s$. Circular padding generally performs better than zero padding, with the largest gain observed on Stanford2D3D, where 2D mIoU improves by 5.22 for the one block configuration. Increasing the number of residual convolutional blocks from one to two further improves performance with circular padding, with the largest gain observed on ToF-360, where 2D mIoU improves by 1.79. The final residual convolutional unit has mixed effects across datasets, but the full configuration, Setting F, provides strong overall performance across the evaluated benchmarks.

\section{Experimental Results}

\subsection{Shared Semantic Categories}
For zero-shot evaluation, different models are trained on different datasets and therefore support different sets of semantic categories. To ensure a fair comparison, we evaluate only the categories that are present in both the training categories of all compared models and the target datasets, Stanford2D3D~\cite{armeni2017joint}, and ToF-360~\cite{kanayama2025tof}. Specifically, for Stanford2D3D, we use 7 shared categories: \texttt{ceiling}, \texttt{chair}, \texttt{door}, \texttt{floor}, \texttt{sofa}, \texttt{table}, and \texttt{wall}. For ToF-360, we use 9 shared categories: \texttt{box}, \texttt{cabinet}, \texttt{ceiling}, \texttt{chair}, \texttt{door}, \texttt{floor}, \texttt{shelf}, \texttt{table}, and \texttt{wall}.

\subsection{Additional Experimental Results}

\begin{table}[t]
\centering
\small
\setlength{\tabcolsep}{1pt}
\resizebox{\columnwidth}{!}{
\begin{tabular}{llccccc}
\toprule
Method & Input & \#Images & \#Params (M) & Mem. (GB) & Time (s) & 3D mIoU$\uparrow$ \\
\midrule
\textbf{Ours} & ERP image & 8 & 1123.7 & 9.56 & \textbf{1.096} & \textbf{17.28} \\
\midrule
\multirow{2}{*}{PanSt3R}
& Cube & 48 & \multirow{2}{*}{906.3} & \textbf{9.28} & 24.938 & 9.26 \\
& 15 Crops & 120 & & 18.07 & 118.781 & 5.18 \\
\midrule
\multirow{2}{*}{IGGT}
& Cube & 48 & \multirow{2}{*}{1617.3} & 27.67 & 4.638 & 2.30 \\
& 15 Crops & 120 & & OOM & OOM & -- \\
\midrule
\multirow{2}{*}{Uni3R}
& Cube & 48 & \multirow{2}{*}{2754.9} & 26.65 & 4.249 & 1.26 \\
& 15 Crops & 120 & & OOM & OOM & -- \\
\bottomrule
\end{tabular}
}
\caption
{
Efficiency comparison on a scene with eight covisible panoramaic images. We benchmark the number of input images, parameter count, peak memory, model inference time, and 3D mIoU on ScanNet++~\cite{yeshwanthliu2023scannetpp}.
}
\label{tab:efficiency_comparison}
\end{table}

\noindent\textbf{Efficiency Comparison.}
Applying perspective-based models such as PanSt3R~\cite{zust2025panst3r}, IGGT~\cite{li2025iggt}, and Uni3R~\cite{sun2026uni3r} to panoramic inputs requires converting each equirectangular projection (ERP) image into multiple perspective views, increasing the number of inputs processed by the model. We compare this overhead using cubemap projections~\cite{wang2020bifuse} and 15 crops as input configurations for the perspective-based baselines. The 15-crop configuration follows the setting used in our data curation pipeline.. As shown in Table~\ref{tab:efficiency_comparison}, our method processes a scene of eight covisible panoramas directly, whereas the baselines require 48 cube faces or 120 crops. This increased input volume leads to higher inference time and peak memory consumption. In particular, the 15-crop configuration exceeds the 48GB memory capacity of an NVIDIA A6000 for both IGGT and Uni3R, resulting in out-of-memory (OOM) during inference. In contrast, our panorama-native design processes the same scene with only 9.56GB peak memory and 1.10s inference time, while achieving a higher 3D mIoU on ScanNet++. These results demonstrate that PanoSeg3R is more computationally efficient and scalable for multi-view 3D scenes by directly operating on panoramic inputs.

\begin{figure}[!t]
    \centering
    \includegraphics[width=\columnwidth]{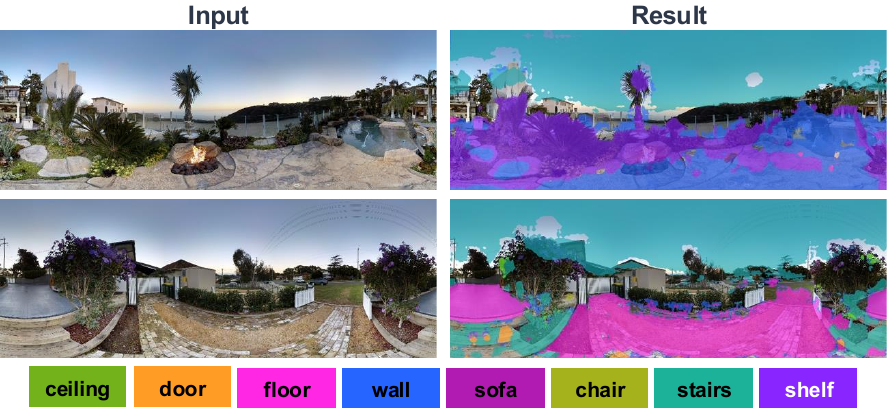}
    \caption
    {
    Failure cases on Matterport3D~\cite{teng2024360bev}. Our model frequently misclassifies or fails to recognize outdoor objects because it is trained primarily on indoor datasets.
    }
    \label{fig:failure_cases}
\end{figure}

\noindent\textbf{Failure Cases.}
Figure~\ref{fig:failure_cases} presents representative failure cases from Matterport3D~\cite{teng2024360bev}. Since our data curation pipeline is constructed using the semantic category set of the indoor ScanNet++~\cite{yeshwanthliu2023scannetpp} dataset, our model primarily learns indoor scene semantics and consequently generalizes poorly to outdoor environments.

\begin{figure*}[t]
    \centering
    \includegraphics[width=\textwidth]{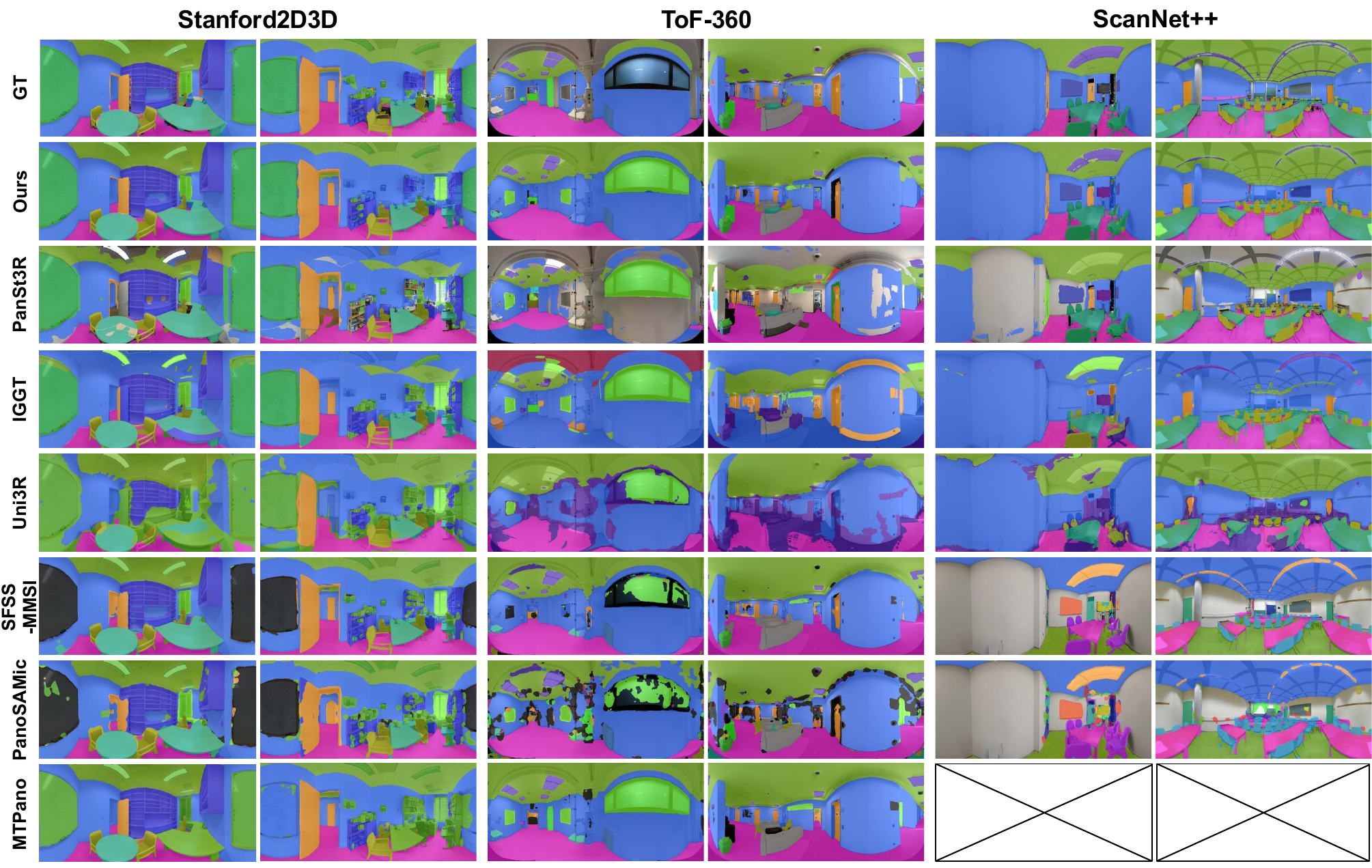}
    \caption
    {
Additional 2D qualitative comparisons with other methods.
    }
    \label{fig:supp_qualitative_comparison2D}
\end{figure*}

\begin{figure*}[t]
    \centering
    \includegraphics[width=\textwidth]{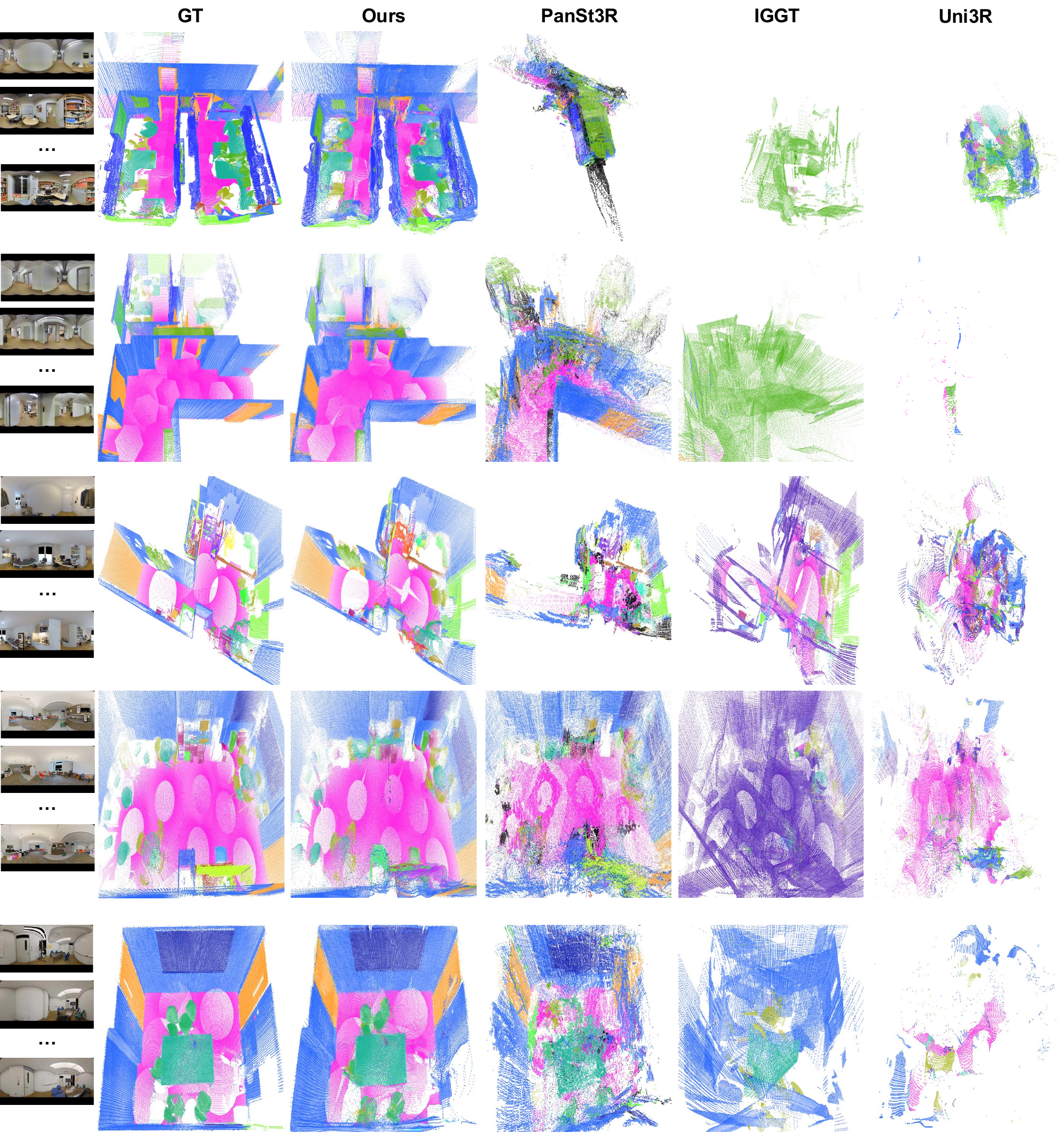}
    \caption
    {
Additional 3D qualitative comparisons with other methods.
    }
    \label{fig:supp_qualitative_comparison3D}
\end{figure*}

\noindent\textbf{Qualitative Result.}
Figure~\ref{fig:supp_qualitative_comparison2D} presents additional 2D qualitative comparisons on Stanford2D3D~\cite{armeni2017joint}, ToF-360~\cite{kanayama2025tof}, and ScanNet++~\cite{yeshwanthliu2023scannetpp}. Figure~\ref{fig:supp_qualitative_comparison3D} shows additional 3D qualitative comparisons, where the first two rows correspond to Stanford2D3D and the remaining three rows correspond to ScanNet++.

\end{document}